\documentclass[vecphys]{svmult}

\usepackage{graphicx}        
\usepackage{multicol}        
\usepackage[bottom]{footmisc}

\makeindex             

\begin{document}

\title*{Optimising the topology of complex neural networks}
\author{Fei Jiang\inst{1,2}
\and Hugues Berry\inst{1}
\and Marc Schoenauer\inst{2}}
\institute{
$^1$Project-team Alchemy and $^2$Project-team TAO, INRIA Futurs, Parc Club Orsay Universite, 91893 Orsay Cedex France,
\texttt{First.Last@inria.fr}}
\maketitle
\begin{abstract}
In this paper, we study instances of complex neural networks, i.e. neural networks with complex topologies. We use Self-Organizing Map neural networks whose neighbourhood relationships are defined by a complex network, to classify handwritten digits. We show that topology has a small impact on performance and robustness to neuron failures, at least at long learning times. Performance may however be increased (by almost $10\%$) by artificial evolution of the network topology. In our experimental conditions, the evolved networks are more random than their parents, but display a more heterogeneous degree distribution.
\end{abstract}

\section{Introduction}

The connectivity structure of complex networks (i.e. their topology) is a crucial determinant of information transfer in large networks (internet, social networks, metabolic networks...)~\cite{Boccal}. Hence, the computation made by complex neural networks, i.e. neural networks with complex connectivity structure, could be dependent on their topology. For instance, recent studies have shown that introducing a small-world topology in a multilayer perceptron increases its performance~\cite{Simard2005, bohland01efficient}. However, other studies have inspected the performance of Hopfield~\cite{Kim04, Morelli04, McGraw03, Stauffer03} or Echo state networks~\cite{DengZhang06} with small-world or scale-free topologies and reported more contrasted results.

Using artificial evolutionary algorithms to modify the topology of neural networks so as to optimise their performance has become widespread in the artificial neural networks community for several years~\cite{NolfiParisi2002,Yao99}. But, in most cases, the studied topologies are quite simple and the number of connections/neurons is low. Furthermore, the evolutionary mechanisms used in most of these studies do not modify the topology in an intensive manner. Hence, the optimisation of large, complex neural networks through artificial evolution has hardly been studied. Note however that in related systems, such has 1D-cellular automata~\cite{sw_a_tgd} or boolean networks~\cite{OAC}, the optimisation of complex topologies has recently begun to confirm the influence of topology on performance and the interest of evolutionary algorithms.

In the case of Self-Organising (or Kohonen) maps (SOMs), the role of network topology has been studied for several years with the perspective of the development of network topologies that preserve that of the data~\cite{Villmann2001}. In the context of complex networks, a different problem is: considering a given data set, do different complex network topologies yield significant differences with respect of performance or robustness?

This paper investigates this issue through an experimental study on the relationship between complex topology and performance for a SOM on a supervised learning problem (handwritten digit classification). The robustness of the results with respect to noise are also addressed. After introducing the context in Section \ref{method}, Section \ref{directProblem} is devoted to the \emph{direct problem}, i.e. observing the performances of networks with different topologies. The \emph{inverse problem} is addressed in Section \ref{inverseProblem}: what topology emerges from the evolutionary optimisation of the classification accuracy of a class of networks?

\section{Methods \& Experiments}
\label{method}

The target of this study are SOMs~\cite{Koho}. The topology/performance relation is looked at from the point of view of recognition/classification of handwritten digits, using the well-known MNIST data base~\cite{MNIST}. SOMs are usually used for unsupervised learning. However, they will be used here for supervised learning, in order to give an unambiguous performance measure. It should be noted that the goal here is not to reach  the best possible performance for the MNIST problem (and indeed SOMs cannot compete with best-to-date published results) but to compare the relative performances of different topologies on the same problem.

Each digit is a $M= 28 \times 28$ pixel picture. The $N$ neurons of the SOM are scattered on a $2d$ space. Each neuron $i$ has an associated $M$-dimensional weight vector $\mathbf{w_i}$, and the different phases of the learning process go as follows (for more details, see, again, \cite{Koho}).\\
\textbf{1. Learning}: At each learning step $t$, a sample digit $\textbf{I}(t)$ is uniformly picked up in the learning dataset. The corresponding Best Matching Unit (BMU) is the neuron whose weight vector is the closest (in $L^2$-norm) to $\textbf{I}(t)$. The weights of the BMU $k$ are updated: $\mathbf{w_k}(t+1) = \mathbf{w_k}(t) + \eta(t)\times(\mathbf{I}(t) - \mathbf{w_k}(t))$. The weights of the neighbours of the BMU are updated similarly, but with a {\em learning rate} $\eta$ that decays following a Gaussian law of the distance with the BMU (the definition of the distance will be given below). Note that the {\em radius} of the neighbourhood (i.e. the variance of the Gaussian law) is decreased along learning iterations.\\
\textbf{2. Labelling}: The basic label of a neuron is the class ($0 \ldots 9$) for which it is the BMU most often over the whole training set. Neurons that never are BMUs are given the basic label of the class from which they are at shortest average distance. In all experiments reported below, the classification label of a neuron is its basic label. Indeed, some tests using as classification label for a neuron some weighted average of the basic labels of its neighbors have shown negligible differences.\\
\textbf{3. Evaluating}: The class given to an unknown example is the label of its BMU. The performance of the network is the misclassification error over a given test set, $F = n_{err}/N_{test}$, where $n_{err}$ is the number of incorrectly classified examples and $N_{test}$ the size of the test set.\\
\noindent \textbf{Distance}: In the classical SOM algorithm, the $N$ neurons are regularly scattered on a $2d$ square grid. The Euclidian distance and the graph distance between neurons (minimum number of hops following an edge) are equivalent. However, when edges are added and suppressed, the situation changes dramatically. Hence, because the goal here is to evaluate the influence of topology on learning performances, the only interesting distance is the graph distance.

\section{Direct problem}
\label{directProblem}
\begin{figure}
\centering
  \includegraphics[width=0.95\textwidth]{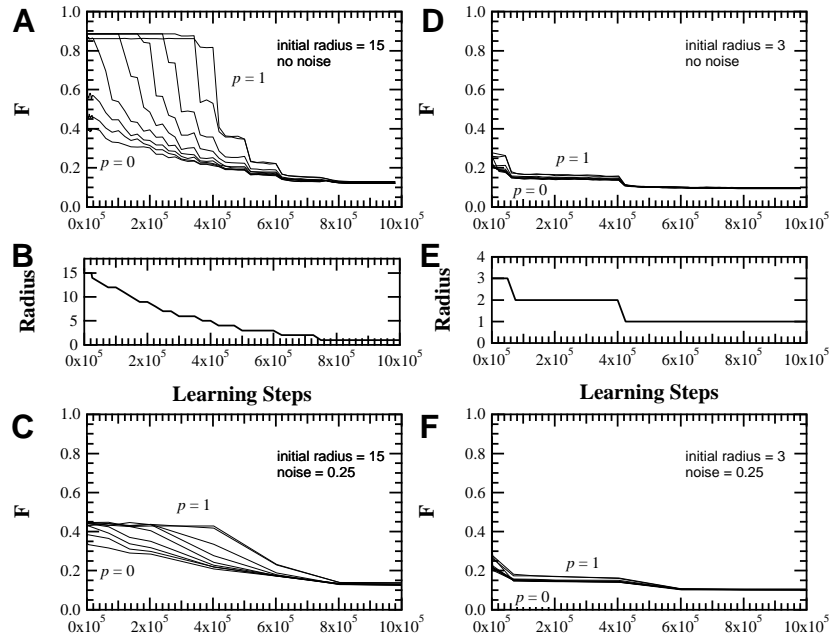}
\caption{\scriptsize Evolution of the performance $F$ during learning for SOMs on complex neighbourhood networks. Neighbourhood networks are constructed positioning neurons on a square grid, and linking each neuron to its 8-nearest neighbours on the grid (Moore neighbourhood). Each link is then rewired to a (uniformly) randomly-chosen destination neuron with probability $p=0, 0.002, 0.004, 0.008, 0.016, 0.032, 0.064, 0.256, 1.000$ (from bottom to top). Panels \emph{A}, \emph{C}, \emph{D} and \emph{F} show the evolution of the fitness $F$ for different values initial radii and noise levels (as indicated on each panels). Panels \emph{B} and \emph{E} display the evolution of the neighbourhood radius. Other parameters: map size $N=1024$ neurons, initial learning rate $\eta(0)=0.080$, training and test sets of 30,000 and 10,000 examples, respectively.}
\label{fig:1}
\end{figure}

\begin{figure}
\centering
\includegraphics[width=0.6\textwidth]{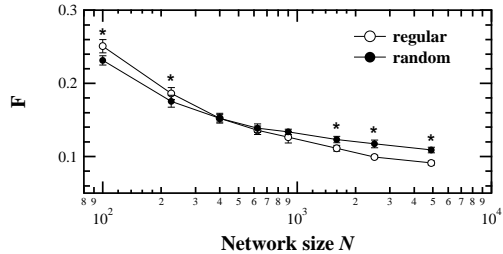}
\caption{\scriptsize Performance $F$ vs number of neurons $N$ after $10^6$ learning steps for (white circles) regular 
or (black circles) uncorrelated random
topologies. Each point is an average over 11 random initial weight and topology realisations. Bars are standard deviations. Stars indicate statistically significant differences (unpaired $t$-test, $p<0.010$). Other parameters as in Fig.~\ref{fig:1}A.}
\label{fig:2}
\end{figure}

The goal of the first experiments was to compare the classification performance of SOM built on the Watts and Strogatz topology model. Figure~\ref{fig:1}A shows the plots of the classification performance $F$ during learning iterations for $N=1024$-neurons networks with regular
to small-world
to fully random
topologies (see caption). This figure shows that, at long learning times, the network performance is clearly not dependent on the topology. This is not surprising since the role of the topology decreases with the neighborhood distance. Important differences are however obvious at short to intermediate times: the more random, the less efficient the network at this time scale. This remark deserves further analysis. Indeed, the performance of these random networks evolves in a piecewise constant fashion. Comparing this evolution to the concomitant decrease of the neighbourhood radius (Fig.~\ref{fig:1}B) uncovers that performance plateaus are synchronised to radius plateaus.\\
The more random the network, the lower its mean shortest path. Hence, a possible interpretation is that, for high $p$ values, the influence of a given neuron at short learning times extends over the entire $2d$ space, to almost every other neuron. Thus, at short time scales, almost all neurons are updated each time a new image is presented, which actually forbids any learning in the network. This interpretation is supported by Fig.~\ref{fig:1}D, where the initial radius is five time smaller than in Fig.~\ref{fig:1}A. Here, the differences in short time behaviours observed above are suppressed.\\
This phenomenon explains a nontrivial effect of noise on the system. Noise is here modeled by deactivating at each learning step a fraction $\nu$ of the neurons (the list of the $N \nu$ deactivated neurons is chosen uniformly for each learning step). Note that all neurons are reactivated for the evaluation phase (step 3 above). Fig.~\ref{fig:1}C shows the performance during learning with $\nu = 0.25$ noise level (i.e. one forth of the neurons are insensitive to learning, at each step) and the same large initial radius than in Fig.~\ref{fig:1}A. Clearly, because the deactivated neurons are protected from update, the above effect of large radius is strongly attenuated. In other words, the presence of noise (here random node failures) actually \emph{improves} the performance of these complex random networks at short learning times. That this effect is effectively related to large radius sizes is confirmed by inspection of Fig.~\ref{fig:1}F, which shows that with small initial radii, this ``beneficial'' effect of noise is not observed (compare with Fig.~\ref{fig:1}D).\\
Another result from Fig.~\ref{fig:1} is that the effects of noise are restricted to short-to-average learning times and disappear with long learning, where the performances of all networks are similar (whatever the topology randomness or initial radius). Hence, at long learning times, the SOMs are robust to neuron failure rates as high as $25\%$, and this robustness does not seem to depend on their neighborhood topology. Finally, Fig.~\ref{fig:2} shows the effects of network size on its performance. While large SOMs ($N > 2,000$) perform better with regular neighborhood networks, the situation is just the opposite with small ($N < 200$) SOMs, where random networks perform better than regular ones. Small-world topologies are intermediate (not shown). Note however that even for the extreme sizes, the difference of fitness between regular and random topologies, though significant, remains minute.\\
%

\section{Inverse problem}
\label{inverseProblem}
The inverse problem consists in optimising the topology in order to minimise the classification error. Evolutionary Algorithms \cite{EibenSmith2003} have been chosen for their flexibility and robustness with respect to local minima. However, due to their high computational cost, only SOMs with $N=100$ neurons could be used.

The algorithm is a mutation-only Steady-State Genetic Algorithm with 2-tournament selection and 6-tournament replacement: At each iteration, the best of two uniformly drawn networks undergoes mutation, and replaces the worse of 6 uniformly drawn networks for the population.

The initial population is composed of 100 different small-world networks ($p=0.050$), and  mutation consists in random rewirings of $C \%$ of uniformly chosen links. $C$ decreases exponentially during evolution ($C=30 \left( 102.6\right) ^{-g/g_{max}}$ where $g$ is the iteration number and $g_{max}$ is the total number of iterations). Here, $g_{max}=200,000$, and $C$ decreases from 102 ($g=0$) downto 1 ($g=g_{max}$). The fitness is computed as the average misclassification error $F$ over 5 learning phases, starting from 5 different initial weights.

\begin{figure}[tb!]
\centering
\includegraphics[width=8cm]{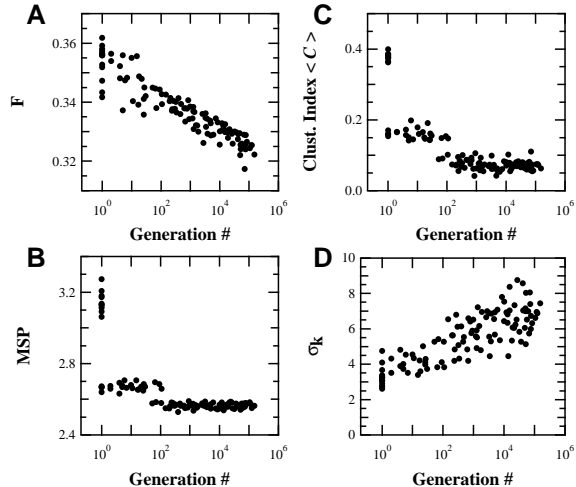} 
\caption{\scriptsize Time courses of the main network statistics during artificial evolution. Each time a mutation gives rise to a topology with a better fitness than the best one in the current population, its fitness (\emph{A}), average mean shortest path (\emph{B}), average clustering index $\left \langle C \right \rangle$ (\emph{C}) and the standard deviation of its connectivity distribution $\sigma_k$ (\emph{D}) are plotted against the current generation number. Each panel groups the results of 9 evolution runs. Parameters: $\eta(0) = 0.35$, fitness estimated as an average over 5 independent runs of 10,000 learning iterations with 2,000 examples from the training set and  5,000 examples from the test set.}
\label{fig:3}
\end{figure}

Considering the small size of these SOMs, one may expect random networks to perform slightly better than regular ones (Fig.~\ref{fig:2}). The main statistics of the best networks obtained during 9 evolution runs are plotted Fig.~\ref{fig:3}. Fig.~\ref{fig:3}A shows that indeed the classification error of the best topology in the population decreases, from $0.355$ to $\approx 0.325$, i.e. a $> 9\%$ improvement.
Interesting characteristics of the best topologies have emerged during evolution: 	  Fig.~\ref{fig:3}B shows an important decrease of the mean shortest path, while Fig.~\ref{fig:3}C demonstrates a clear collapse (more than fourfold reduction) of the clustering index. In other words, the topology evolves towards more randomness -- as could be expected from Fig.~\ref{fig:2}.\\
Interestingly, there is another important change in the topology along evolution, concerning the network connectivity distribution. Indeed, the standard deviation $\sigma_k$ of the connectivity distribution $P(k)$ (where $P(k)$ is the probability that a neuron chosen at random has $k$ neighbours) almost triples during evolution (Fig.~\ref{fig:3}D). This means that the connectivity distribution of the networks broadens (becomes less sharply peaked). In other words, artificial evolution yields more heterogeneous networks. However, it should be clear that this result is highly dependent on the topology of the data themselves (here MNIST datebase), and could be different with other data.

\section{Conclusion}

The objective of this paper was to study the influence of topology in a case of neural network defined on a complex topology. On the limited experiments presented here, it seems that the performance of the network is only weakly controlled by its topology. Though only regular, small-world and random topologies, have been presented, similar results have been obtained for scale-free topologies. This suggests that for such learning task, the topology of the network is not crucial.

Interestingly, though, these slight differences can nevertheless be exploited by evolutionary algorithms: After evolution, the networks are more random than the initial small-world topology population. Their connectivity distribution is also more heterogeneous, which may indicate a tendency to evolve toward scale-free topologies. Unfortunately, this assumption can only be tested with large-size networks, for which the shape of the connectivity distribution can unambiguously be determined, but whose artificial evolution, for computation cost reasons, could not be carried out. Similarly, future work will have to address other classical computation problems for neural networks before we are able to draw any general conclusion.
%
%
\footnotesize


\end{document}